\documentclass[conference]{IEEEtran}
\usepackage{epsfig,graphicx,parskip,setspace,tabularx,xspace}
\usepackage[caption=false,font=footnotesize]{subfig}
\usepackage{booktabs}

\ifCLASSINFOpdf
\else
\fi
\usepackage[cmex10]{amsmath}
\usepackage{amssymb}
\newcommand{\argmax}{\operatornamewithlimits{argmax}}

\begin{document}
%
\title{X-CNN: Cross-modal Convolutional Neural Networks for Sparse Datasets}

\author{\IEEEauthorblockN{Petar Veli\v{c}kovi\'{c}\IEEEauthorrefmark{1}\IEEEauthorrefmark{3}, Duo Wang\IEEEauthorrefmark{1}, Nicholas D. Lane\IEEEauthorrefmark{2}\IEEEauthorrefmark{3} and Pietro Li\`{o}\IEEEauthorrefmark{1}}
\IEEEauthorblockA{\IEEEauthorrefmark{1}Computer Laboratory, University of Cambridge, Cambridge CB3 0FD, UK\\
\IEEEauthorrefmark{2}Department of Computer Science, University College London, London
WC1E 6BT, UK\\
\IEEEauthorrefmark{3}Nokia Bell Labs, Cambridge CB3 0FA, UK\\
Email: \{pv273, wd263, pl219\}@cam.ac.uk, niclane@acm.org}}



\maketitle

\begin{abstract}
In this paper we propose \emph{cross-modal convolutional neural networks} (X-CNNs), a novel biologically inspired type of CNN architectures, treating gradient descent-specialised CNNs as individual units of processing in a larger-scale network topology, while allowing for unconstrained \emph{information flow} and/or \emph{weight sharing} between analogous hidden layers of the network---thus generalising the already well-established concept of neural network \emph{ensembles} (where information typically may flow only between the output layers of the individual networks). The constituent networks are individually designed to learn the output function on their own subset of the input data, after which cross-connections between them are introduced after each pooling operation to periodically allow for information exchange between them. This injection of knowledge into a model (by prior partition of the input data through domain knowledge or unsupervised methods) is expected to yield greatest returns in sparse data environments, which are typically less suitable for training CNNs. For evaluation purposes, we have compared a standard four-layer CNN as well as a sophisticated FitNet4 architecture against their cross-modal variants on the CIFAR-10 and CIFAR-100 datasets with differing percentages of the training data being removed, and find that at lower levels of data availability, the X-CNNs significantly outperform their baselines (typically providing a 2--6\% benefit, depending on the dataset size and whether data augmentation is used), while still maintaining an edge on \emph{all} of the full dataset tests.
\end{abstract}


%
\IEEEpeerreviewmaketitle

\section{Introduction}

In recent years, the number of success stories of machine learning has seen an all-time rise across a wide range of fields and tasks, examples including: computer vision \cite{AlexNet}, speech recognition \cite{hinton2012deep}, reinforcement learning \cite{mnih2015human} and guiding Monte Carlo tree search \cite{silver2016mastering}. The unifying idea behind all of the above is \emph{deep learning}, the utilisation of neural networks with many hidden layers, for the purposes of learning complex feature representations from raw data, rather than relying on hand-crafted feature extraction. 

As the networks become deeper, however, they become more and more reliant on the amount of training examples provided for maximising their performance. While we are now able to extract large quantities of labelled information for many problems of interest, there remains a significant proportion of tasks for which ``big data'' simply isn't available at this time, which makes it extremely difficult to fully exploit a deep CNN architecture and properly learn generalisable features of the data. Here we will present an architectural methodology that attempts to extract additional predictive power from a convolutional neural network (CNN) in such circumstances by instead focussing on the \emph{width} of the data, i.e. the heterogeneity of information present within each training example. The key idea constitutes appropriate \emph{partitioning} of this information and training \emph{smaller CNNs} on these partitions (allowing them to train faster and more effectively under sparse data environments), while allowing for information exchange between them at various stages (Fig. \ref{figmcnn}). 

A classic example where such an approach is bound to be useful are \emph{clinical studies}, where there typically may not be that many patients, but for each patient there is potentially a heterogeneous wealth of information, such as various test results, patient history, ethnic background, body scans (CT, MRI\dots) and so on, depending on the type of study.

\section{Cross-modal CNNs}

Our methodology is inspired by \emph{multilayer networks} \cite{kivela2014multilayer}, mathematical structures encompassing several layers of graphs over the same set of nodes, allowing for unrestricted intra-layer as well as inter-layer connections. They have been a demonstrably valuable tool for modelling a variety of natural and social systems (\cite{de2016physics,estrada2014communicability,granell2014competing}), and their applicability to machine learning (within the context of hidden Markov models) was already demonstrated by some of the authors \cite{Veličković27122015}, managing to achieve high performance on a sparse breast cancer classification dataset involving gene expression and methylation data.

\begin{figure*}
	\centering
	\includegraphics[width=\linewidth]{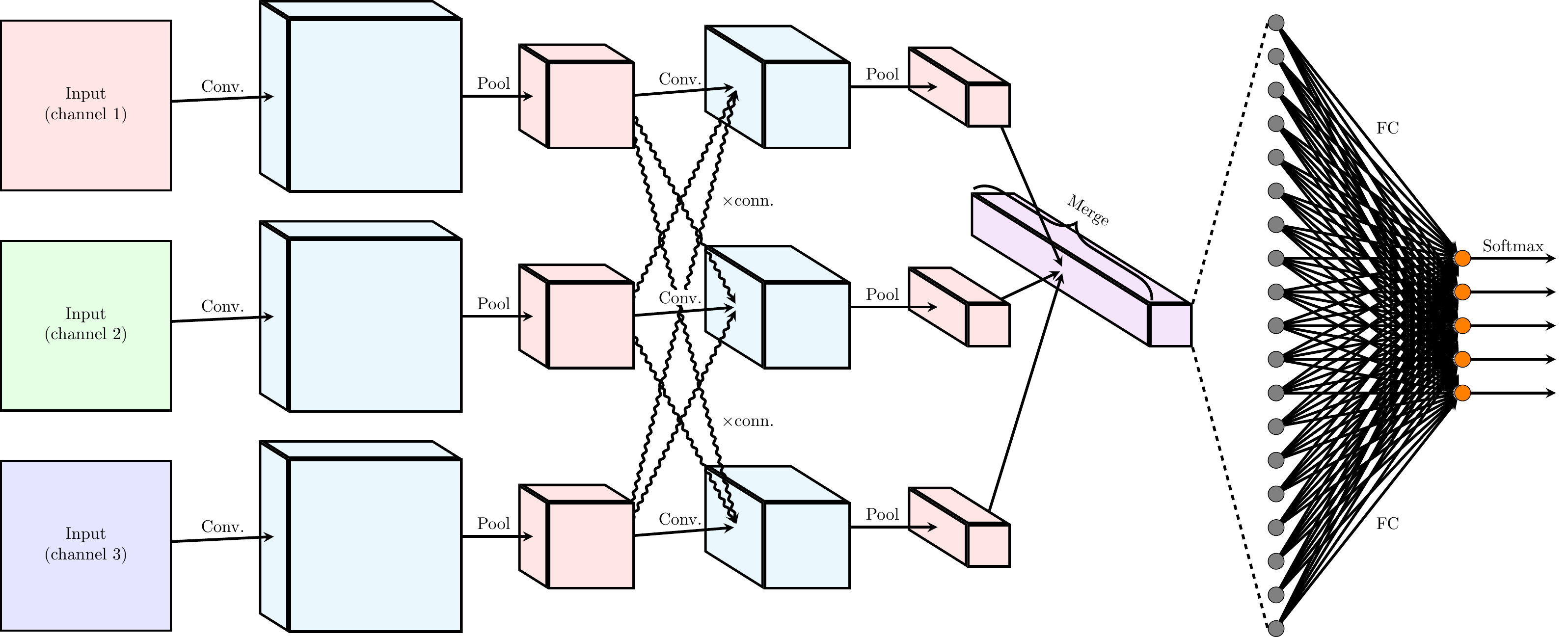}
    \caption{Diagram of a simple cross-modal CNN for image classification, generated from a baseline CNN of the form $\left[Conv \rightarrow Pool\right] \times 2 \rightarrow FC \rightarrow Softmax$. Each of the three channels (RGB/YUV) of the input image receives its own CNN superlayer, with cross-connections inserted after the pooling operation, and full weight sharing in the fully connected layers. A more in-depth view of a potential cross-connection layout is provided by Figure \ref{figxtalk}.}
    \label{figmcnn}
\end{figure*}

The network design process is initiated by appropriately partitioning the input data---this may be done either manually (by exploiting existing domain knowledge) or through an unsupervised pre-training step, which will determine which (not necessarily disjoint) fragments of the input data are more likely to constructively influence one another. Afterwards, a cross-modal CNN is constructed such that a separate CNN \emph{superlayer} is dedicated to each partition of the input data, attempting to learn the target function from its partition only.  The purpose of the partitioning is to help the constituent CNNs become powerful predictors while requiring a smaller dimensionality of the input data, by allowing them access to those parts of the input which are most significantly related to each other in the context of the predictions that need to be made.

Finally, the superlayers may be interconnected by any sort of (feedforward) cross-connection as is best seen fit, and they may be combined in arbitrary ways at the output stage to produce the final output. Similarly, at any stage the weights of the superlayers may be shared---the simplest case, which we will explore in our analysis, constitutes complete weight sharing of the fully connected layers at the tail of the networks. This construction is biologically inspired by \emph{cross-modal systems} \cite{eckert2008cross} within the visual and auditory systems of the human brain (which in turn inspired the development of CNNs)---wherein several cross-connections between various sensory networks have been discovered \cite{Beer2011,yang2015}.

\begin{figure*}
	\centering
	\includegraphics[width=\linewidth]{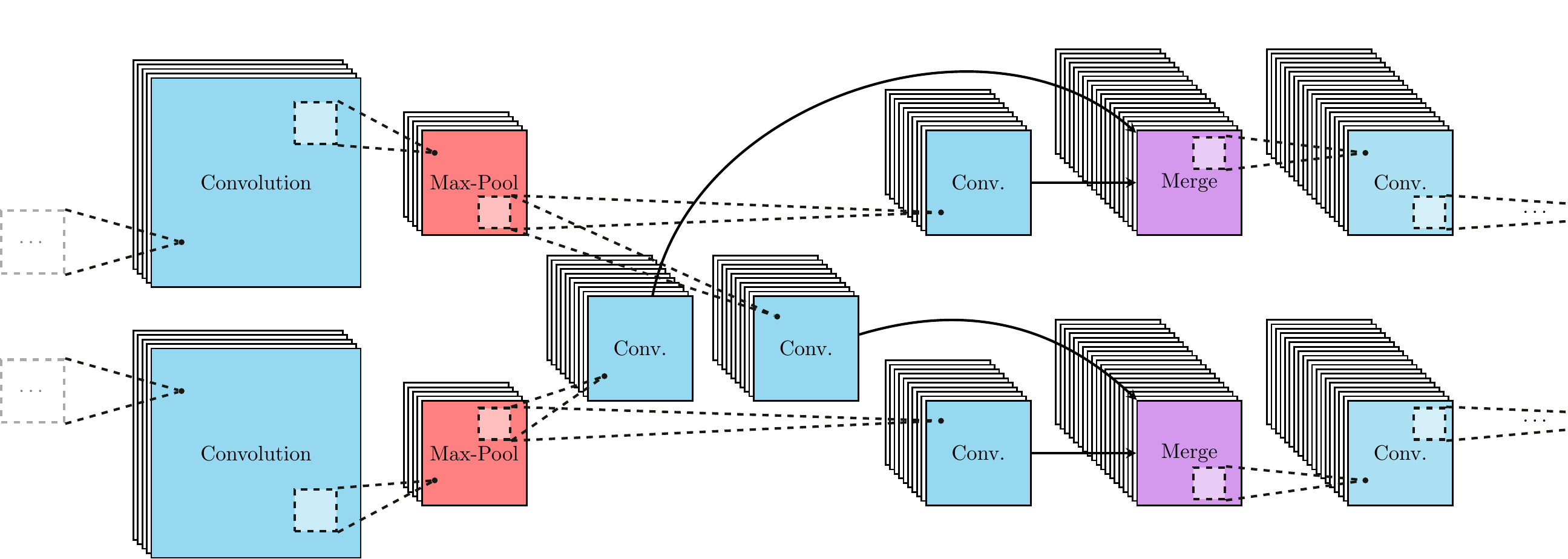}
    \caption{Illustration of a single cross-connection segment within an X-CNN with two superlayers. After each pooling operation, we exchange the feature maps between the superlayers, after first passing them through an additional convolutional layer. We may also perform an additional intra-superlayer convolution before merging the feature maps in each superlayer via concatenation.}
    \label{figxtalk}
\end{figure*}

To quantify the gains of this approach, our evaluation focusses on an already well-understood problem of coloured image classification, on established CIFAR-10/100 \cite{krizhevsky2009learning} benchmarks for which an abundance of data is available, so it is easier to investigate the effects of restricting the size of the training set on various CNN models. The partitioning of the input that we consider is per-channel---each of the three image channels will be an input to an individual superlayer, and these superlayers will have identical high-level architecture (differing only in the number of feature maps per hidden layer)---as illustrated by Fig. \ref{figmcnn}. This also allows for a simple approach to cross-connections; namely, after every downsampling (\emph{pooling}) operation we allow for the feature maps to be exchanged between superlayers, after being passed through another convolutional layer (Fig. \ref{figxtalk}).

While this model in itself constitutes a committee of CNNs, it differs from most traditional ensemble applications in two key ways: 
\begin{itemize}
	\item An ensemble's constituent models typically exchange information only in the \emph{output} stage, while the cross-modal framework allows for arbitrary (feedforward) information flow at any stage of the processing pipeline;
	\item Constituent models of an ensemble usually receive a full copy of the input each, while superlayers within a cross-modal neural network receive only a fraction of the input, allowing for a \emph{decrease in degrees-of-freedom} of the model compared to an unrestricted network.
\end{itemize}
In fact, this can be taken a step further: one may consider \emph{ensembles of cross-modal CNNs}, which may compound on benefits already given by X-CNNs themselves, on examples where the networks are potentially struggling to choose a proper class with sufficient confidence. As the X-CNN model can be observed as an ordinary CNN from a high level, \emph{any} ensemble strategies that are found useful for CNNs should be useful for X-CNNs as well.

Lastly, it should be noted that our approach is not restricted to CNNs, but it is then easiest to scrutinise, as the trained parameters are bound to obey a certain spatial structure. In line with this, an entire section of this manuscript will be dedicated to analysing the learned convolutional kernels within an X-CNN, as well as visualising the inputs that would maximise the model's cross-connection activations.

\section{Model Architectures}

For the purposes of evaluating our proposed architecture's performance, we have implemented two baseline CNN models---along with their cross-modal variants---in Keras \cite{chollet2015keras} (with Theano \cite{2016arXiv160502688short} back-end). For purposes of reproducibility, in this section we will expose their architectures and hyperparameters as used for the evaluation. The cross-modal variants' feature map counts have been altered in such a way as to make the overall number of parameters as close as possible to the baseline, making for fair evaluation with respect to degrees-of-freedom.

For both of the models used, we represent images in the YUV colour space. As a linear transformation from RGB, it should not have an impact on performance of the baselines, while it has the benefit of decoupling \emph{luminance} from \emph{chrominance}, allowing for a simpler analysis of cross-connections (and relating its learned kernels to human vision processes). We inject further domain knowledge into the model by favouring the CNN superlayer corresponding to the Y channel in terms of feature map counts (typically doubled compared to the U/V superlayers within the same hidden layer). This corresponds to the assumption that the majority of relevant information about an object is contained within its brightness channel, while colour usually represents auxiliary information.

\subsection{KerasNet}

Our initial model of choice represents a simple CNN with four convolutional ReLU \cite{nair2010rectified} layers, followed by two fully connected layers, one of which is also ReLU. We will be referring to it as KerasNet throughout this manuscript as it is based on the Keras CIFAR-10 CNN example \cite{chollet2015kerasnet}. It represents a likely style of a ``starting'' model that one is going to attempt to apply on an image classification problem (without particular prior knowledge about it), perhaps especially bearing in mind that the training data may be sparse.

The architecture of the model, as well as its cross-modal variant (X-KerasNet) is outlined in Table \ref{tblkerasnet}. Both models are trained for 200 epochs using the Adam SGD optimiser, with hyperparameters as described in \cite{kingma2014adam}, and a batch size of 32. Dropout \cite{srivastava2014dropout} has been applied after both of the pooling operations (with $p = 0.25$) as well as after the first fully connected layer (with $p = 0.5$). 

\begin{table}
    \caption{Architectures for KerasNet and X-KerasNet}
    \centering
	\begin{tabular}{c c c} \toprule
    	Output size & KerasNet & X-KerasNet\\ 
        & $\sim4.46$M param. & $\sim4.37$M param.\\ \midrule
        $32 \times 32$ & $[3 \times 3, 64] \times 2$ & Y: $[3 \times 3, 32] \times 2$\\
        & & U/V: $[3 \times 3, 16] \times 2$\\
        $16 \times 16$ & \multicolumn{2}{c}{$2\times 2$ Max-Pool, stride $2$} \\
        & & Y $\rightarrow$ Y: \emph{identity}\\
        & & U $\rightarrow $ U: \emph{identity}\\
        & & V $\rightarrow $ V: \emph{identity}\\
        & & Y $\rightsquigarrow $ U/V: $[1 \times 1, 32]$\\
        & & U/V $\rightsquigarrow $ Y: $[1 \times 1, 16]$\\
        & $[3 \times 3, 128] \times 2$ & Y: $[3 \times 3, 64] \times 2$\\
        & & U/V: $[3 \times 3, 32] \times 2$\\
        $8 \times 8$ & \multicolumn{2}{c}{$2\times 2$ Max-Pool, stride $2$} \\
        $1 \times 1$ & \multicolumn{2}{c}{Fully connected, 512-D} \\ 
        & \multicolumn{2}{c}{10/100-way softmax}\\
        \bottomrule
    \end{tabular}    
    \label{tblkerasnet}
\end{table}

\subsection{FitNet4}

We decided to implement FitNet4 by Romero \emph{et al.} \cite{romero2014fitnets} as our second baseline, representing a sophisticated CNN close to the state-of-the-art on CIFAR-10/100. We opted for this model as it is prominently featured in a variety of recent neural networks research (\cite{srivastava2015training,mishkin2015all}), and due to its design goal of being a ``thin\&deep'' network, managing to keep its parameter count relatively low compared to many other successful models, and therefore could still be a feasible first choice for handling a sparse dataset.

The FitNet4 consists of 17 convolutional 2-way maxout \cite{goodfellow2013maxout} layers, followed by two fully connected layers, the first of which is a 5-way maxout layer. The full architecture of this model---as well as its cross-modal variant (X-FitNet4)---is presented in Table \ref{tblfitnet}.

\begin{table}
    \caption{Architectures for FitNet4 and X-FitNet4}
    \centering
	\begin{tabular}{c c c} \toprule
    	Output size & FitNet4 & X-FitNet4\\ 
        & $\sim2.75$M param. & $\sim2.72$M param.\\ \midrule
        $32 \times 32$ & $[3 \times 3, 32] \times 3$ & Y: $[3 \times 3, 24] \times 3$\\
        & & U/V: $[3 \times 3, 12] \times 3$\\
        & $[3 \times 3, 48] \times 2$ & Y: $[3 \times 3, 36] \times 2$\\
        & & U/V: $[3 \times 3, 18] \times 2$\\
        $16 \times 16$ & \multicolumn{2}{c}{$2\times 2$ Max-Pool, stride $2$} \\
        & & Y $\rightarrow$ Y: $[1 \times 1, 36]$\\
        & & U $\rightarrow $ U: $[1 \times 1, 18]$\\
        & & V $\rightarrow $ V: $[1 \times 1, 18]$\\
        & & Y $\rightsquigarrow $ U/V: $[1 \times 1, 12]$\\
        & & U/V $\rightsquigarrow $ Y: $[1 \times 1, 12]$\\
        & $[3 \times 3, 80] \times 6$ & Y: $[3 \times 3, 60] \times 6$\\
        & & U/V: $[3 \times 3, 30] \times 6$\\
        $8 \times 8$ & \multicolumn{2}{c}{$2\times 2$ Max-Pool, stride $2$} \\
        & & Y $\rightarrow$ Y: $[1 \times 1, 60]$\\
        & & U $\rightarrow $ U: $[1 \times 1, 30]$\\
        & & V $\rightarrow $ V: $[1 \times 1, 30]$\\
        & & Y $\rightsquigarrow $ U/V: $[1 \times 1, 18]$\\
        & & U/V $\rightsquigarrow $ Y: $[1 \times 1, 18]$\\
        & $[3 \times 3, 128] \times 6$ & Y: $[3 \times 3, 96] \times 6$\\
        & & U/V: $[3 \times 3, 48] \times 6$\\
        $1 \times 1$ & \multicolumn{2}{c}{$8 \times 8$ (global) Max-Pool} \\
        & \multicolumn{2}{c}{Fully connected, 500-D} \\ 
        & \multicolumn{2}{c}{10/100-way softmax}\\
        \bottomrule
    \end{tabular}    
    \label{tblfitnet}
\end{table}

Both models are initialised using Xavier initialisation \cite{glorot2010understanding}, and are then trained for 230 epochs using the Adam SGD optimiser with a batch size of 128. We have applied batch normalisation \cite{ioffe2015batch} to the output of each hidden layer to significantly accelerate the training procedure. $L_2$ regularisation with $\lambda = 0.0005$ has been applied to all weights in the model. Finally, dropout (with $p=0.2$) was applied on the input, after every pooling operation, and after the fully connected maxout layer.

\section{Evaluation}\label{sect}

To verify our insights, we have utilised two well-known image classification benchmark datasets, CIFAR-10 and CIFAR-100 \cite{krizhevsky2009learning}, for which an abundance of data is available (50000 training and 10000 testing examples). This makes it easier to study the behaviour of the considered CNNs as different fractions of the training data are discarded. We hypothesise that, at lower levels of data availability (up to a threshold), our methodology will yield significant gains over an equivalent unrestricted CNN---and also that it will remain competitive at all higher training set sizes. 

The validity of our claim is investigated by performing comparative evaluation, with the KerasNet and FitNet4 as baselines against X-KerasNet and X-FitNet4, respectively. In each individual test we evaluate the accuracy of these four models on the entire test set of 10000 samples, when the training routine is presented with only $p\%$ of the entire training dataset (chosen deterministically). The schedule for the tests is as follows:

\begin{itemize}
	\item Initially test in increments of $5\%$, until reaching $20\%$ (at which time the training and testing sets have equal sizes);
    \item Afterwards, test in increments of $10\%$ until either reaching $50\%$ or the accuracies of the two models get within $0.5\%$ of each other (corresponding to a gain of $\leq 50$ images properly classified), whichever is later;
    \item Specially, we always test on $1\%$ (corresponding to a highly sparse environment with only $500$ training images) and $100\%$ of the training dataset.
\end{itemize}

The images are preprocessed by applying a single batch normalisation operation on them; we have found this to yield slightly better results compared to doing global contrast normalisation and ZCA whitening (the more common approach). Finally, given that it is, depending on the task, sometimes possible to significantly enhance results in a sparse environment by way of \emph{data augmentation}, we have run all of the above tests \emph{twice}---with and without random translations and horizontal reflections applied to the training images---providing insight as to whether data augmentation compounds the effects of a cross-modal architecture, and to what extent.

\section{Results and Discussion}

The full evaluation results on the aforementioned tests are presented in Tables \ref{tblcif10noaug}--\ref{tblcif100aug}.

\begin{table*}
	\setlength\tabcolsep{1.5pt}
	\centering
	\caption{Comparative Evaluation Results on CIFAR-10 Without Data Augmentation}
	\begin{tabular}{l r r r r r r r r r r r r r} \toprule
    \raisebox{-0.3em}{\text{\normalsize\bf Model}}\ $\big\backslash$\ \raisebox{0.3em}{\text{\emph{\textbf{p}}}} & \multicolumn{1}{c}{1\%} & \multicolumn{1}{c}{5\%} & \multicolumn{1}{c}{10\%} & \multicolumn{1}{c}{15\%} & \multicolumn{1}{c}{20\%} & \multicolumn{1}{c}{30\%} & \multicolumn{1}{c}{40\%} & \multicolumn{1}{c}{50\%} & \multicolumn{1}{c}{60\%} & \multicolumn{1}{c}{70\%} & \multicolumn{1}{c}{80\%} & \multicolumn{1}{c}{90\%} & \multicolumn{1}{c}{100\%} \\ \midrule
    KerasNet & 37.94\% & 53.82\% & 62.95\% & 67.39\% & 70.26\% & 74.39\% & 76.62\% & 78.55\% & --------- & --------- & --------- & --------- & 82.50\%\\
    X-KerasNet & \bf 41.19\% & \bf 57.84\% & \bf 65.01\% & \bf 68.25\% & \bf 71.36\% & \bf 74.79\% & \bf 76.96\% & \bf 78.57\%  & --------- & --------- & --------- & --------- & \bf 82.62\%\\
    FitNet4 & 38.97\% & 56.78\% & 70.37\% & 75.07\% & 78.50\% & 81.95\% & 83.95\% & 85.22\% & --------- & --------- & --------- & --------- & 89.56\%\\
    X-FitNet4 & \bf 39.21\% & \bf 60.57\% & \bf 70.82\% & \bf 76.09\% & \bf 79.40\% & \bf 83.36\% & \bf 84.25\% & \bf 86.14\% & --------- & --------- & --------- & --------- & \bf 90.13\%\\ \bottomrule
    \end{tabular}
    \label{tblcif10noaug}
\end{table*}
\begin{table*}
	\setlength\tabcolsep{1.5pt}
	\centering
	\caption{Comparative Evaluation Results on CIFAR-10 With Data Augmentation}
	\begin{tabular}{l r r r r r r r r r r r r r} \toprule
    \raisebox{-0.3em}{\text{\normalsize\bf Model}}\ $\big\backslash$\ \raisebox{0.3em}{\text{\emph{\textbf{p}}}} & \multicolumn{1}{c}{1\%} & \multicolumn{1}{c}{5\%} & \multicolumn{1}{c}{10\%} & \multicolumn{1}{c}{15\%} & \multicolumn{1}{c}{20\%} & \multicolumn{1}{c}{30\%} & \multicolumn{1}{c}{40\%} & \multicolumn{1}{c}{50\%} & \multicolumn{1}{c}{60\%} & \multicolumn{1}{c}{70\%} & \multicolumn{1}{c}{80\%} & \multicolumn{1}{c}{90\%} & \multicolumn{1}{c}{100\%} \\ \midrule
    KerasNet & 45.45\% & 67.01\% & 70.89\% & 78.83\% & \bf 80.97\% & \bf 83.23\% & 83.64\% & 85.02\% & --------- & --------- & --------- & --------- & 86.66\%\\
    X-KerasNet & \bf 49.60\% & \bf 69.28\% & \bf 72.51\% & \bf 78.96\% & 80.58\% & 83.10\% & \bf 83.89\% & \bf 85.37\% & --------- & --------- & --------- & --------- & \bf 87.41\%\\
    FitNet4 & 40.91\% & \bf 65.73\% & 75.55\% & 80.85\% & 83.63\% & 86.23\% & \bf 88.30\% & 89.11\% & --------- & --------- & --------- & --------- & 92.27\%\\
    X-FitNet4 & \bf 42.02\% & 65.54\% & \bf 77.06\% & \bf 81.33\% & \bf 83.94\% & \bf 86.41\% & 88.13\% & \bf 89.37\% & --------- & --------- & --------- & --------- & \bf 92.50\%\\ \bottomrule
    \end{tabular}
    \label{tblcif10aug}
\end{table*}
\begin{table*}
	\setlength\tabcolsep{1.5pt}
	\centering
	\caption{Comparative Evaluation Results on CIFAR-100 Without Data Augmentation}
	\begin{tabular}{l r r r r r r r r r r r r r} \toprule
    \raisebox{-0.3em}{\text{\normalsize\bf Model}}\ $\big\backslash$\ \raisebox{0.3em}{\text{\emph{\textbf{p}}}} & \multicolumn{1}{c}{1\%} & \multicolumn{1}{c}{5\%} & \multicolumn{1}{c}{10\%} & \multicolumn{1}{c}{15\%} & \multicolumn{1}{c}{20\%} & \multicolumn{1}{c}{30\%} & \multicolumn{1}{c}{40\%} & \multicolumn{1}{c}{50\%} & \multicolumn{1}{c}{60\%} & \multicolumn{1}{c}{70\%} & \multicolumn{1}{c}{80\%} & \multicolumn{1}{c}{90\%} & \multicolumn{1}{c}{100\%} \\ \midrule
    KerasNet & 7.55\% & 15.10\% & 20.24\% & 24.76\% & 28.18\% & 32.43\% & 36.29\% & 38.61\% & 41.63\% & 44.10\% & 45.56\% & 46.26\% & 48.26\%\\
    X-KerasNet & \bf 8.05\% & \bf 16.45\% & \bf 23.04\% & \bf 26.91\% & \bf 30.08\% & \bf 35.39\% & \bf 39.13\% & \bf 41.88\% & \bf 42.50\% & \bf 45.96\% & \bf 46.73\% & \bf 48.25\% & \bf 49.98\%\\
    FitNet4 & 6.48\% & 16.84\% & 22.12\% & 28.30\% & 35.52\% & 39.28\% & 43.59\% & 49.69\% & 50.42\% & 55.83\% & 56.62\% & 58.00\% & 59.78\%\\
    X-FitNet4 & \bf 6.64\% & \bf 18.73\% & \bf 27.57\% & \bf 33.59\% & \bf 38.38\% & \bf 45.53\% & \bf 49.68\% & \bf 52.21\% & \bf 55.55\% & \bf 57.22\% & \bf 59.52\% & \bf 60.87\% & \bf 62.20\%\\ \bottomrule
    \end{tabular}
    \label{tblcif100noaug}
\end{table*}
\begin{table*}
	\setlength\tabcolsep{1.5pt}
	\centering
	\caption{Comparative Evaluation Results on CIFAR-100 With Data Augmentation}
	\begin{tabular}{l r r r r r r r r r r r r r} \toprule
    \raisebox{-0.3em}{\text{\normalsize\bf Model}}\ $\big\backslash$\ \raisebox{0.3em}{\text{\emph{\textbf{p}}}} & \multicolumn{1}{c}{1\%} & \multicolumn{1}{c}{5\%} & \multicolumn{1}{c}{10\%} & \multicolumn{1}{c}{15\%} & \multicolumn{1}{c}{20\%} & \multicolumn{1}{c}{30\%} & \multicolumn{1}{c}{40\%} & \multicolumn{1}{c}{50\%} & \multicolumn{1}{c}{60\%} & \multicolumn{1}{c}{70\%} & \multicolumn{1}{c}{80\%} & \multicolumn{1}{c}{90\%} & \multicolumn{1}{c}{100\%} \\ \midrule
    KerasNet & 9.09\% & 24.68\% & 32.63\% & 38.64\% & 42.62\% & 47.64\% & 49.91\% & 52.46\% & 53.77\% & 54.26\% & 55.12\% & 55.42\% & 55.45\%\\
    X-KerasNet & \bf 10.16\% & \bf 27.15\% & \bf 35.58\% & \bf 42.05\% & \bf 43.77\% & \bf 48.80\% & \bf 50.48\% & \bf 54.25\% & \bf 54.90\% & \bf 55.33\% & \bf 55.68\% & \bf 56.82\% & \bf 57.18\%\\
    FitNet4 & 7.25\% & 17.94\% & 23.55\% & 29.24\% & 38.76\% & 48.07\% & 50.06\% & 56.01\% & 58.55\% & 59.80\% & 62.38\% & 63.60\% & 65.59\%\\
    X-FitNet4 & \bf 7.35\% & \bf 20.39\% & \bf 28.69\% & \bf 37.86\% & \bf 43.75\% & \bf 50.48\% & \bf 55.40\% & \bf 57.92\% & \bf 60.70\% & \bf 62.76\% & \bf 66.18\% & \bf 66.27\% & \bf 67.19\%\\ \bottomrule
    \end{tabular}
    \label{tblcif100aug}
\end{table*}

\begin{figure*}[!t]
	\centering
  \captionsetup[subfigure]{labelformat=empty}
  \subfloat[][]{\includegraphics[width=0.4\linewidth]{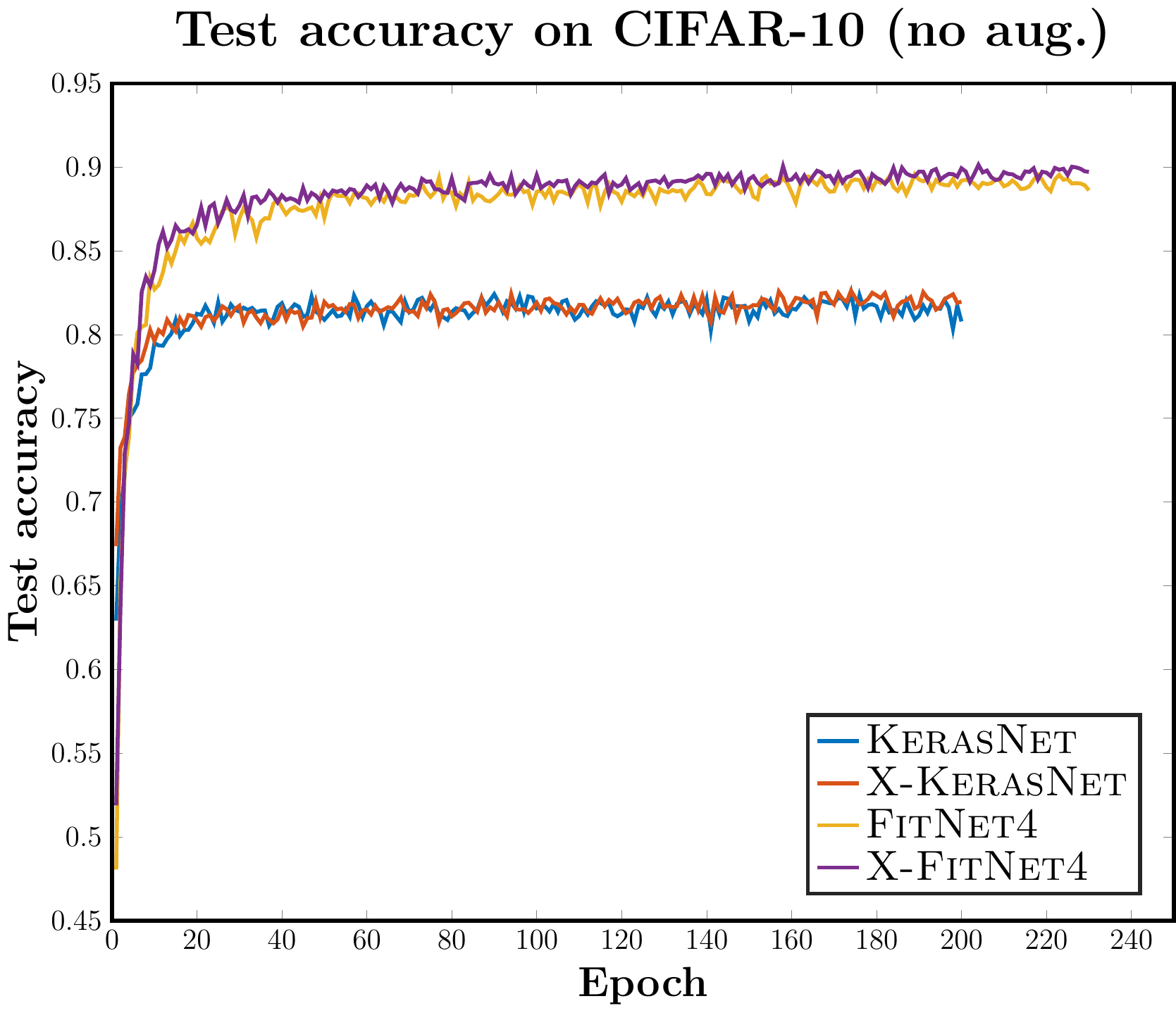}}\quad
  \subfloat[][]{\includegraphics[width=0.4\linewidth]{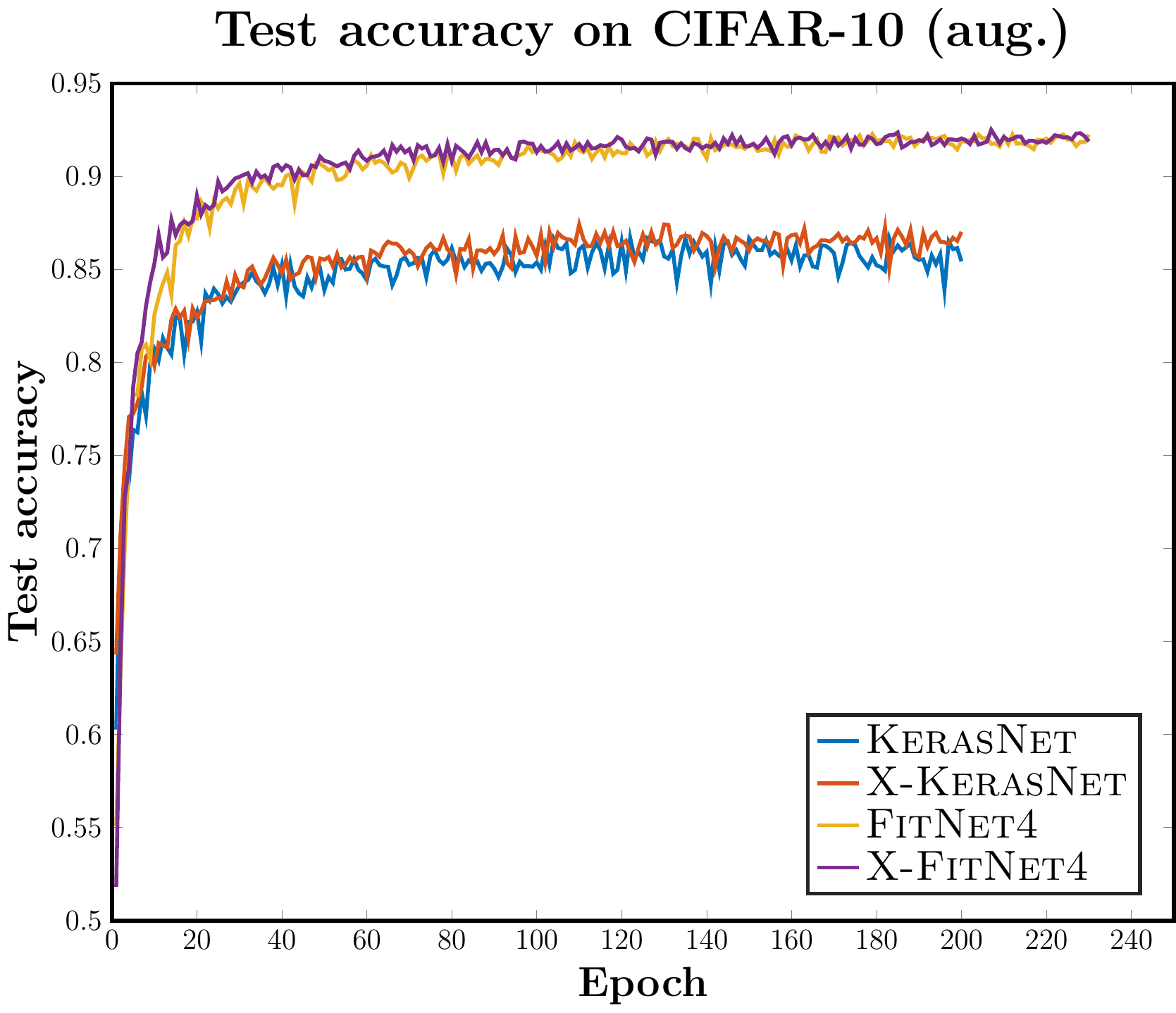}}\\[-4ex]
  \subfloat[][]
  {\includegraphics[width=0.4\linewidth]{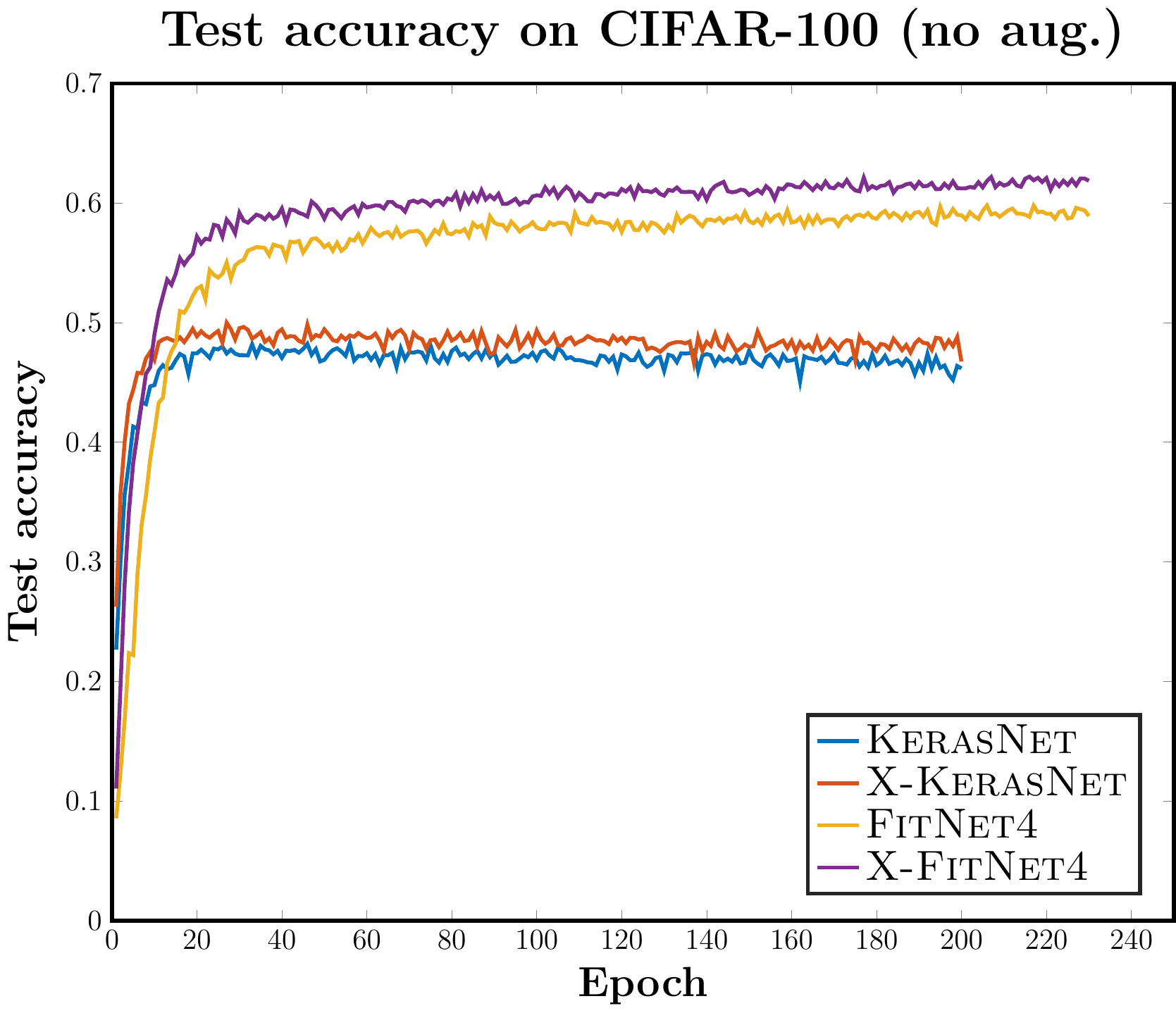}}\quad
  \subfloat[][]{\includegraphics[width=0.4\linewidth]{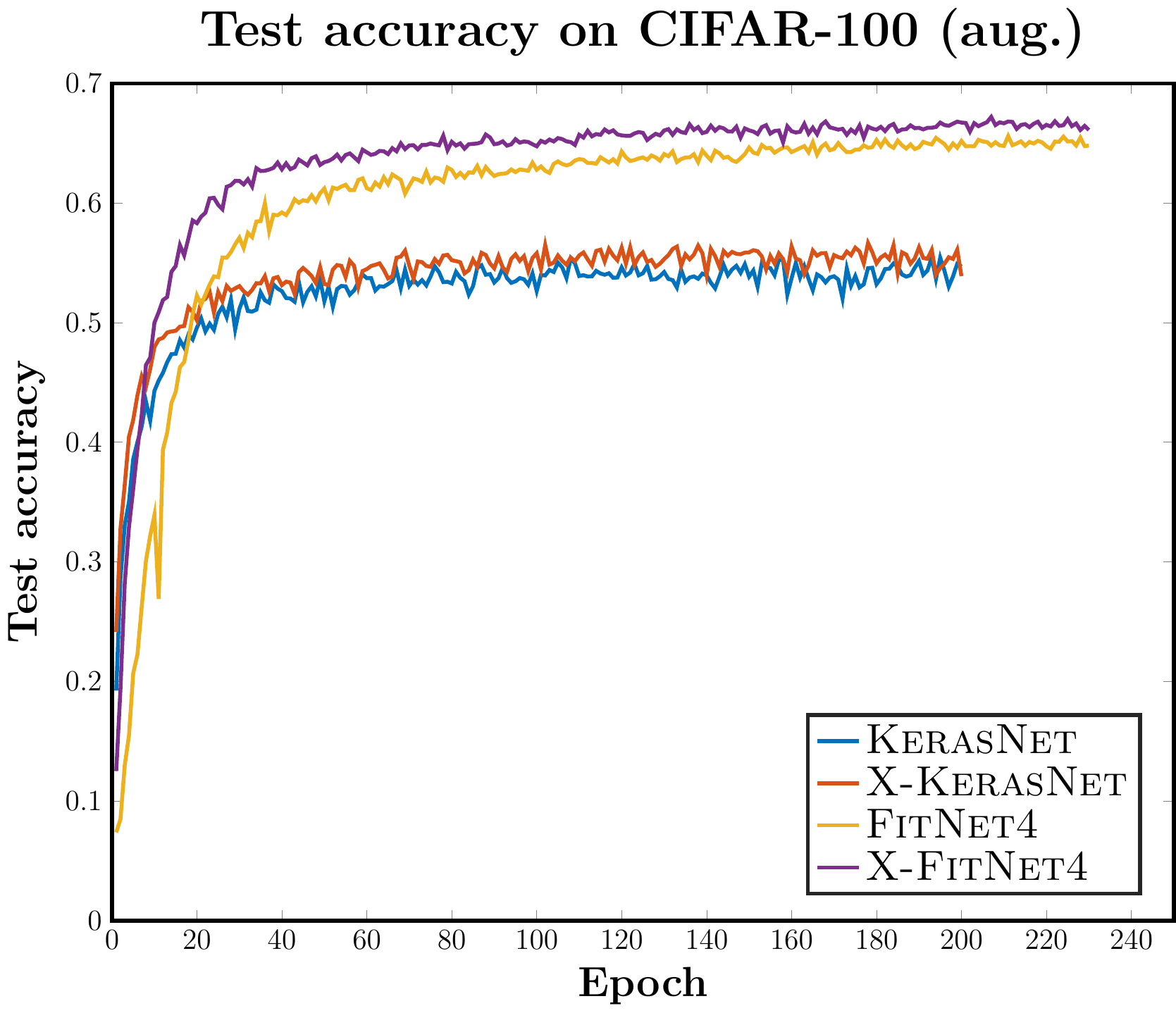}}
	\caption{Plots of the test accuracy of the four CNN models under consideration as a function of the number of training epochs, under 100\% of the training set available. The experiments have been carried out on both CIFAR-10 and CIFAR-100, with and without data augmentation. The cross-modal CNNs are consistently competitive with their respective baselines across all four datasets, with a significant edge present for CIFAR-100.}
    \label{figplots}
\end{figure*}

The results on tests without data augmentation are completely in line with the claim of Section \ref{sect}; at sufficiently low training data sizes, both X-KerasNet and X-FitNet4 \emph{significantly outperform} their respective baselines on the testing set, for both of the CIFAR-10/100 datasets.

For CIFAR-10, the threshold at which the baselines ``catch up'' (in terms of being able to manually learn the domain knowledge directly injected into their cross-modal variants) is at around $p = 40\%$, corresponding to 20000 training examples being available. Furthermore, on CIFAR-100, such a threshold is \emph{never} reached, most likely due to the extreme sparsity of per-class examples making this problem particularly suitable for the X-CNN models; the only exception is the $1\%$ scenario for FitNet4, where the data sparsity is probably too extreme (five examples/class) for such a deep model to reach its potential.

Regardless of when the threshold is surpassed, we report that the cross-modal CNNs will generally continue to have a slight edge over the baselines---outperforming them on \emph{all} of the full training dataset experiments, sometimes significantly. This naturally invites the conclusion that converting a CNN into a X-CNN (if allowed by the task) is always a reasonable step; it can yield significant benefits (the significance depending on the relation between the sparsity of the training dataset and the complexity of the baseline model), while rarely making performance significantly worse.

To further verify this claim, we have performed experiments on the full datasets (with and without augmentation) where we monitored how the testing accuracy evolves as a function of training epoch. The resulting plots are summarised in Figure \ref{figplots}; it is clear that the X-CNNs are at least as powerful as their baselines, even when the full training sets are available. Furthermore, it is possible to detect a narrow edge for the cross-modal models in the CIFAR-10 experiments, and a significant edge in the CIFAR-100 experiments. The concluding remark is that even when the dataset under investigation is not very sparse, attempting to utilise a cross-modal variant of the considered models (if applicable) is a reasonable action, as it might yield noticeable returns in predictive power.

The analysis of the interplay between data augmentation and cross-modal networks on CIFAR-100 remains straightforward---the X-CNN models remaining consistently and significantly ahead of their baselines throughout the entire spectrum of training set sizes. On CIFAR-10, however, this is slightly more complicated; while the catch-up threshold gets expectedly decreased (to around $p=20\%$), the behaviour of X-CNNs for smaller training set sizes does not always significantly compound the benefits of data augmentation. Specifically, at 5\% of the training set the FitNet4 model manages to outperform X-FitNet4 (the roles do get reversed starting from 10\%, however). As a possible cause of this phenomenon, we note that, at this data availability level, both of the FitNet4 models are significantly inferior in performance to the KerasNet models, for which there is a significant benefit to the usage of X-CNNs. The takeaway lesson here is that, while the cross-modal architecture need not always compound nicely with data augmentation, an occurrence of such an event could signify that the baseline was not particularly suitable for properly accommodating data augmentation at this training set size in the first place. If this happens, one should attempt to use a more suitable/shallower CNN---the X-CNN variant should then produce the desired benefits.

Finally, we have taken advantage of some of the smaller training set sizes to perform \emph{statistical significance tests}, typically scarce in deep learning literature. For training set sizes up to 15\%, we trained the models five times (from different initial conditions) and then performed $t$-tests, choosing $p < 0.05$ as our significance threshold. Our findings show that, under these assumptions, the best-performing X-CNN model's performance advantages are \emph{statistically significant in all scenarios}, aside from the data-augmented CIFAR-10.

\section{Cross-connection Analysis}
A key element of the X-CNN architecture are the cross-connection layers, as they enable information flow between individual channels. It will therefore be of interest to understand and visualise what is the mode of operation for these layers. All of the visualisations in this section correspond to the learned weights after fully training on 100\% of CIFAR-10 with data augmentation.

\begin{figure}
  \captionsetup[subfigure]{labelformat=empty}
  \centering
  \subfloat[]{\includegraphics[width = 0.71\linewidth]{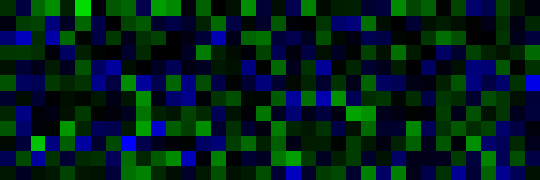}}\\[-4.5ex]

  \subfloat[]{\includegraphics[width = 0.35\linewidth]{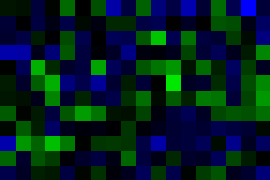}}\ 
  \subfloat[]{\includegraphics[width = 0.35\linewidth]{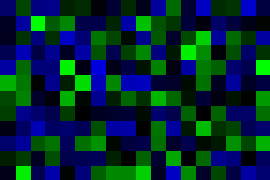}}


  \caption{Weight visualisation of the first-level cross-connection layer for the X-FitNet4 CNN. The columns correspond to input channels, while rows correspond to output channels. Green colour indicates a positive-weight connection between an input channel and an output channel, while blue colour indicates a negative-weight connection. The colour intensities are proportional to the absolute weight values. Top: Y $\rightsquigarrow$ U/V (36 input channels,
  12 output channels). Bottom, left-to-right: U $\rightsquigarrow$ Y and V $\rightsquigarrow$ Y (18 input channels, 12 output channels).}
  \label{fig:weight_visualization}
  \end{figure}

We will first demonstrate that cross-connections inserted in the considered models, though being $1\times1$ convolutions, learn more complex functions than simple feature map passing. First, we note that the weights of a $1\times1$ convolutional layer may be represented as a 2D table that maps input channels to output channels (akin to an adjacency matrix, where columns are the input channels and rows are the output channels). Rather than displaying the raw table values, we decided to visualise weights in a \emph{heatmap} style; Figure \ref{fig:weight_visualization} showcases this visualisation for the first cross-connection layer of X-FitNet4. Green colours indicate that an input channel has a positive connection weight to the respective output channel while blue colours indicate negative weights. The colour intensities are proportional to the absolute weight values.

It can be seen that each output channel of the cross-connection layer is obtained through a \emph{nontrivial weighted combination} of input channels. We hypothesise that the cross-connection layers selectively filter and combine input features that are more utilisable in another processing stream.

To delve deeper into what kinds of features the cross-connection layers are filtering, combining and passing, we applied layer-wise 
feature-map activation techniques proposed by Simonyan \emph{et al.} \cite{simonyan2013deep}. This technique performs gradient ascent on a white-noise input image to maximise activations of a specific channel of feature maps at any of the layers within a pre-trained model. The objective function for gradient ascent is defined as
\begin{equation} \label{eq:vis}
 {\bf I'} = \argmax_{\bf I} \Sigma({\bf I})-\lambda\|{\bf I}\|^2
\end{equation}
where ${\bf I}$ is input image, $\Sigma({\bf I})$ is the activation of the considered neuron when provided with ${\bf I}$ as input, and $\lambda$ is a regularisation factor. After iterating for a number of gradient ascent steps, the original white-noise image will be modified into patterns that approximate the detection function of a specific neuron.

\begin{figure}
  \captionsetup[subfigure]{labelformat=empty}
  \centering
  \subfloat[]{\includegraphics[width = 0.25\linewidth]{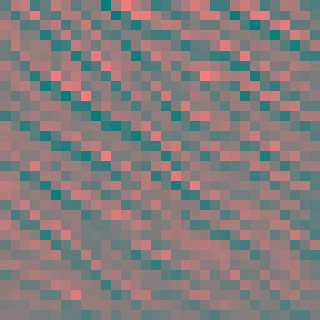}}\ 
  \subfloat[]{\includegraphics[width = 0.25\linewidth]{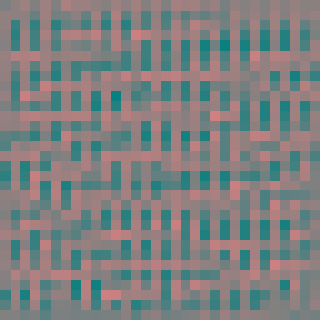}}\ 
  \subfloat[]{\includegraphics[width = 0.25\linewidth]{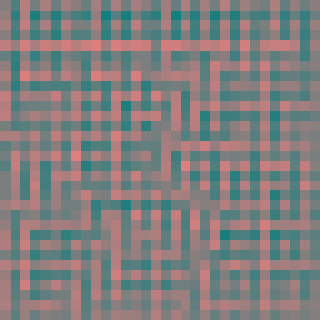}}\\[-4.5ex]
  
  \subfloat[]{\includegraphics[width = 0.25\linewidth]{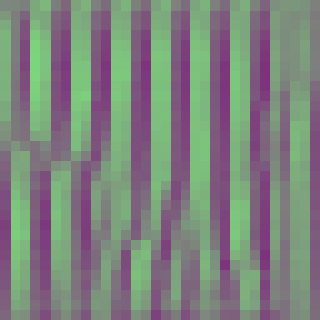}}\ 
  \subfloat[]{\includegraphics[width = 0.25\linewidth]{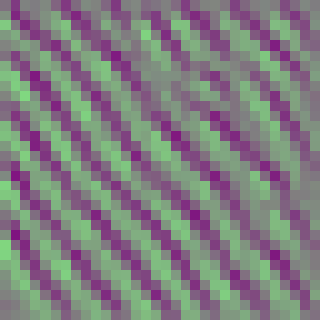}}\ 
  \subfloat[]{\includegraphics[width = 0.25\linewidth]{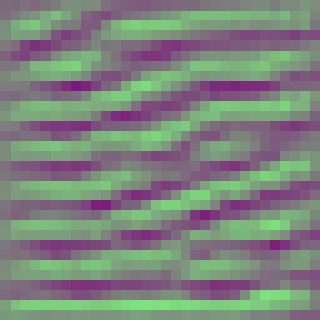}}\\[-4.5ex]
  
  \subfloat[]{\includegraphics[width = 0.25\linewidth]{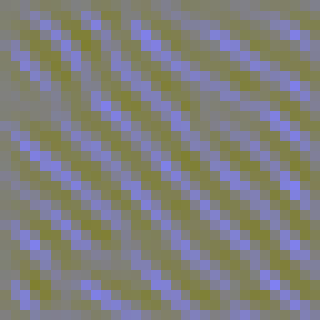}}\ 
  \subfloat[]{\includegraphics[width = 0.25\linewidth]{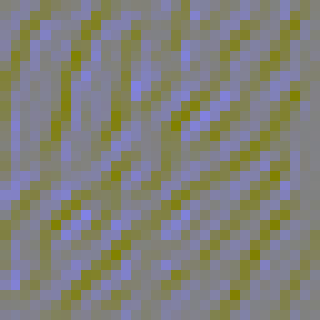}}\ 
  \subfloat[]{\includegraphics[width = 0.25\linewidth]{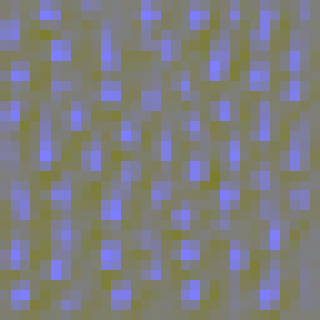}}

  \caption{Artificially generated images (from white noise) that cause strong activations of specific channels in the first cross-connection layer of the X-FitNet4 model. Top: Three channels from the Y $\rightsquigarrow$ U/V cross-connections. Middle: Three channels from the U $\rightsquigarrow$ Y cross-connections. Bottom: Three channels from the V $\rightsquigarrow$ Y cross-connections.}
  \label{fig:layer_visualization}
  \end{figure}

Lower-level convolutional layers are well-known to learn filters approximating Gabor wavelet filters that act as edge detectors, corner detectors, etc; we can confirm that in our experiments this has indeed been the case. For the first cross-connection layer of X-FitNet4, we have visualised a selection of channel activations in Figure \ref{fig:layer_visualization}. This visualisation indicates that the cross-connection layer is indeed 
passing combined lower-level features, such as the addition of horizontal and vertical stripes in the upper right image in the figure. We observe further that the pattern frequency for the Y channel's crossconnection layer is higher than the one for the U and V layers. This observation reflects the fact that the human vision system is able to detect higher frequency variations in intensity than chrominance. This is a solid indicator that the X-CNN architecture, when faced with an image classification task in the YUV colour scheme, is actually attempting to mimic human vision.
  
\begin{figure}
  \captionsetup[subfigure]{labelformat=empty}
  \centering
  \subfloat[]{\includegraphics[width = 0.2\linewidth]{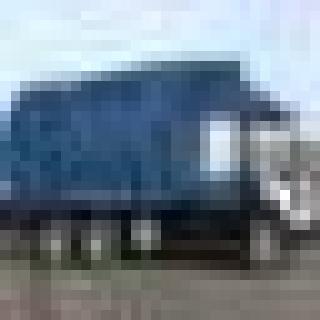}}\hspace{0.1em}
  \subfloat[]{\includegraphics[width = 0.2\linewidth]{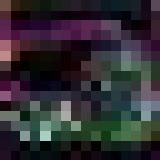}}\hspace{0.1em}
  \subfloat[]{\includegraphics[width = 0.2\linewidth]{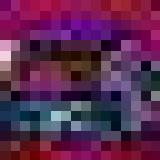}}\\[-4ex]
  
  \subfloat[]{\includegraphics[width = 0.2\linewidth]{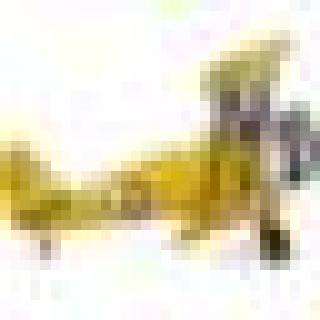}}\hspace{0.1em}
  \subfloat[]{\includegraphics[width = 0.2\linewidth]{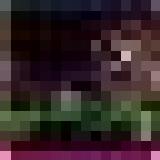}}\hspace{0.1em}
  \subfloat[]{\includegraphics[width = 0.2\linewidth]{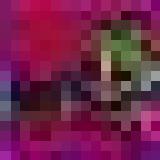}}
  

  \caption{Visualisation of input and output feature maps of the Y $\rightsquigarrow$ U/V cross-connection layer of X-KerasNet. Left: input images (\emph{truck}/\emph{airplane}). Middle: input feature maps to the cross-connection layer. Right: output feature maps of the cross-connection layer.}
  \label{fig:feature_map}
  \end{figure}

Our final analysis focusses on the X-KerasNet model, where we transformed feature maps of arbitrary depths into RGB images by a colour-mapping scheme. Figure \ref{fig:feature_map} shows the feature maps of the inputs and outputs of the Y $\rightsquigarrow$ U/V cross connections for representative images of the \emph{truck} and \emph{airplane} classes. These were easier to comparatively analyse on the X-KerasNet, as its cross-connection layers do not alter the number of feature maps, and therefore the same colour-mapping scheme remained meaningful for both. We observe that cross-connection output maps have background and some features emphasised, while other features de-emphasised---which further indicates that the cross-connection layers are performing more complex inter-superlayer feature integration than simply passing feature maps between superlayers.

\section{Conclusion}

We have introduced cross-modal convolutional neural networks (X-CNNs), a novel architecture that decouples convolutional processing of (typically image-based) input partitions, while allowing for periodical information flow between the processing pipelines, in order to achieve performance improvements in sparse data environments. We have applied this methodology on the popular CIFAR-10/100 image classification datasets for two baseline models, managing to significantly outperform them in low-data environments, while remaining competitive in high-data environments---outperforming them on \emph{all} of the full-dataset experiments.

Aside from reinforcing the claim that the X-CNN architecture can only be beneficial to a baseline model (depending on the levels of training data sparsity, potentially highly significantly), we have further verified that the introduced cross-connection layers perform rather complex functions (thus they are not limited to simple feature map passing) and are capable of mimicking human vision processes---confirming that the biological inspiration behind such a model is justified.






\bibliographystyle{IEEEtran}
\bibliography{refs}
%



\end{document}